\documentclass{article}

\usepackage{microtype}
\usepackage{graphicx}
\usepackage{booktabs} 

\usepackage{amsmath,amssymb,amsfonts}
\usepackage{algorithmic}
\usepackage{graphicx}
\usepackage{textcomp}
\usepackage{xcolor}
\usepackage{enumitem}
\usepackage{xfrac}
\usepackage{color}
\usepackage{subcaption}
\usepackage{listings}
\usepackage{multirow}
\usepackage{bigdelim}
\usepackage{setspace}
\usepackage{calc}
\usepackage{fdsymbol}
\usepackage{xfrac}
\usepackage{soul}
\usepackage{hhline}
\usepackage[T1]{fontenc}

\definecolor{lightyellow}{RGB}{250, 250, 180}
\definecolor{HLYELLOW}{RGB}{240, 127, 0}
\definecolor{hlyellow}{RGB}{240, 127, 0}
\sethlcolor{lightyellow}

\newcommand{\tenscons}{$\medtriangleup$}
\newcommand{\statnorm}{\makebox[\widthof{\tenscons}][c]{$\largesquare$}}
\newcommand{\elewise}{\makebox[\widthof{\tenscons}][c]{$\medcircle$}}
\newcommand{\atttenscons}{$\medblacktriangleup$}
\newcommand{\attstatnorm}{\makebox[\widthof{\tenscons}][c]{$\largeblacksquare$}}
\newcommand{\attelewise}{\makebox[\widthof{\tenscons}][c]{$\medblackcircle$}}
\newcommand{\opsymbspace}{\makebox[\widthof{\tenscons}][c]{}}
\newcommand{\mue}{\textrm{MUE}}

\newcommand{\mr}[2]{\multirow{#1}{*}{#2}}

\setlist[itemize]{leftmargin=*,nosep}
\setlist[enumerate]{leftmargin=*,nosep}
\setlist[description]{leftmargin=1em,nosep}
\setlist{nolistsep}

\PassOptionsToPackage{hyphens}{url}\usepackage{hyperref}

\usepackage{hyperref}

\usepackage[accepted]{mlsys2021}

\mlsystitlerunning{Data Movement Is All You Need}

\begin{document}

\twocolumn[
\mlsystitle{Data Movement Is All You Need: A Case Study on Optimizing Transformers}



\mlsyssetsymbol{equal}{*}

\begin{mlsysauthorlist}
\mlsysauthor{Andrei Ivanov}{equal,eth}
\mlsysauthor{Nikoli Dryden}{equal,eth}
\mlsysauthor{Tal Ben-Nun}{eth}
\mlsysauthor{Shigang Li}{eth}
\mlsysauthor{Torsten Hoefler}{eth}
\end{mlsysauthorlist}

\mlsysaffiliation{eth}{Department of Computer Science, ETH Z\"{u}rich, Switzerland}

\mlsyscorrespondingauthor{Andrei Ivanov}{anivanov@inf.ethz.ch}

\mlsyskeywords{Data movement, high-performance computing, deep learning, transformers}

\vskip 0.3in

\begin{abstract}
Transformers are one of the most important machine learning workloads today. Training one is a very compute-intensive task, often taking days or weeks, and significant attention has been given to optimizing transformers. Despite this, existing implementations do not efficiently utilize GPUs. We find that data movement is the key bottleneck when training. Due to Amdahl's Law and massive improvements in compute performance, training has now become memory-bound. Further, existing frameworks use suboptimal data layouts. Using these insights, we present a recipe for globally optimizing data movement in transformers. We reduce data movement by up to 22.91\% and overall achieve a 1.30$\times$ performance improvement over state-of-the-art frameworks when training a BERT encoder layer and 1.19$\times$ for the entire BERT. Our approach is applicable more broadly to optimizing deep neural networks, and offers insight into how to tackle emerging performance bottlenecks.
\end{abstract}
]



\printAffiliationsAndNotice{\mlsysEqualContribution} 

\section{Introduction}
\label{sec:intro}
\begin{figure*}[ht!]
    \centering
    \begin{minipage}[b]{0.5\textwidth}
        \begin{subfigure}[b]{\linewidth}
\begin{lstlisting}[
            	basicstyle=\fontsize{7}{9}\ttfamily\bfseries,
            	language=Python,
            	alsoletter={.},
            	tabsize=4,
            	stringstyle=\color{red!50!black},
            	emph={q,k,v,bq,bk,bv,bo,wq,wk,wv,wo,scaler,qq,kk,vv,beta,alpha,gamma,out},
            	emphstyle={\color{violet!50!black}},
            	emph={[2]mha_forward, dropout,softmax, np.einsum, dace.float16},emphstyle={[2]\color{blue!50!black}},
            	morekeywords={@dace.program}
            ]
@dace.program
def mha_forward(
  wq: dace.float16[P, H, I], bq: dace.float16[P, H],
  q: dace.float16[I, B, J],
  wk: dace.float16[P, H, I], bk: dace.float16[P, H],
  k: dace.float16[I, B, K],
  wv: dace.float16[W, H, I], bv: dace.float16[W, H],
  v: dace.float16[I, B, K],
  wo: dace.float16[W, H, I], bo: dace.float16[I],
  scaler: dace.float16
):
  qq = np.einsum("phi,ibj->phbj", wq, q) + bq[:,:,None,None]
  kk = np.einsum("phi,ibk->phbk", wk, k) + bk[:,:,None,None]
  vv = np.einsum("whi,ibk->whbk", wv, v) + bv[:,:,None,None]
  beta = scaler * np.einsum("phbk,phbj->hbjk", kk, qq)
  alpha = dropout(softmax(beta))
  gamma = np.einsum("whbk,hbjk->whbj", vv, alpha)
  out = np.einsum("whi,whbj->ibj", wo, gamma)+bo[:,None,None]
  return out
\end{lstlisting}
            \caption{Input Code}
            \label{fig:sdfg-code}
        \end{subfigure}
    \end{minipage}
    \quad
     \begin{subfigure}[b]{0.45\textwidth}
        \includegraphics[width=1.\linewidth]{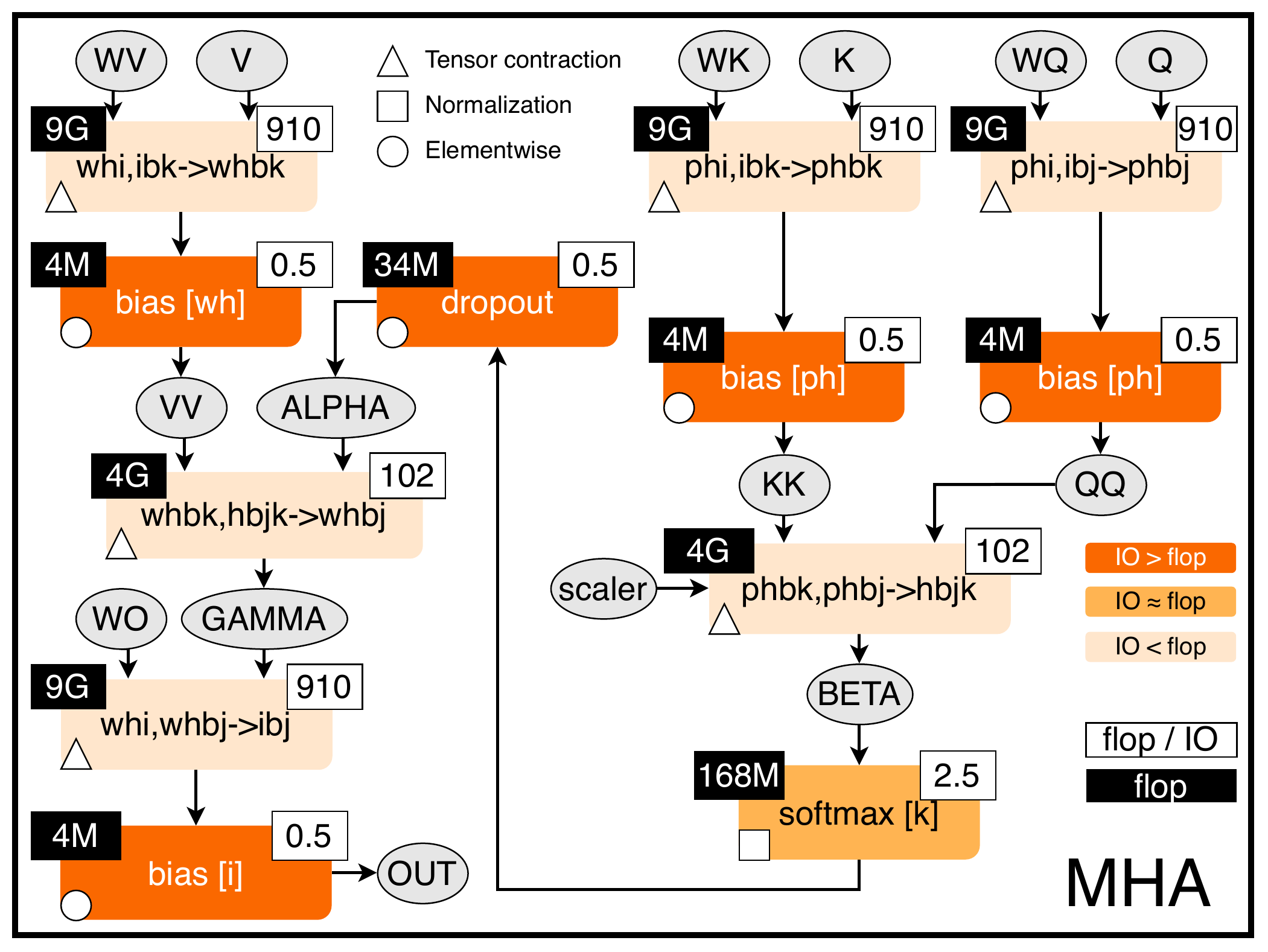}
        \caption{Resulting dataflow}
        \label{fig:mha-sdfg}
    \end{subfigure}
    \caption{Input code and stateful dataflow multigraph (SDFG) for Multi-Head Attention. Throughout the paper, if not stated otherwise, the values are given for the following set of parameters: $P = W = 64$, $H = 16$, $I = P \cdot H = 1024$, $B = 8$, $J = K = 512$. P/W: key/value projection size; H: \# heads; I: embedding size; B: batch size; J/K: 
    input/output sequence length.}
    \vspace{-1em}
    \label{fig:mha}
\end{figure*}

Transformers~\cite{vaswani2017attention} are a class of deep neural network architecture for sequence transduction~\cite{graves2012sequence}, with similar applicability as RNNs~\cite{rumelhart1986learning} and LSTMs~\cite{hochreiter1997long}. They have recently had a major impact on natural language processing (NLP), including language modeling~\cite{radford2018improving,wang2018glue,wang2019superglue}, question-answering~\cite{rajpurkar2018know}, translation~\cite{vaswani2017attention}, and many other applications. The significant improvement in accuracy brought by transformers to NLP tasks is comparable to the improvement brought to computer vision by AlexNet~\cite{krizhevsky2012imagenet} and subsequent convolutional neural networks. Transformers have also begun to be applied to domains beyond NLP, including vision~\cite{dosovitskiy2020image}, speech recognition~\cite{yeh2019transformer}, reinforcement learning~\cite{parisotto2019stabilizing}, molecular property prediction~\cite{maziarka2020molecule}, and symbolic mathematics~\cite{lample2019deep}.

Training transformers is very compute-intensive, often taking days on hundreds of GPUs or TPUs~\cite{devlin2018bert,yang2019xlnet,liu2019roberta,keskar2019ctrl,shoeybi2019megatron,lan2019albert,raffel2019exploring}. Further, generalization only improves with model size~\cite{radford2019language,shoeybi2019megatron,raffel2019exploring,microsoft2020turingnlg}, number of training samples~\cite{raffel2019exploring,liu2019roberta}, and total iterations~\cite{liu2019roberta,kaplan2020scaling}. These all significantly increase compute requirements. Indeed, transformers are becoming the dominant task for machine learning compute where training a model can cost tens of thousands to millions of dollars and even cause environmental concerns~\cite{strubell2019energy}. These trends will only accelerate with pushes toward models with tens of billions to trillions of parameters~\cite{microsoft2020turingnlg,microsoft2020zero}, their corresponding compute requirements~\cite{openai2018aiandcompute}, and increasing corporate investment towards challenges such as artificial general intelligence~\cite{openai2019microsoft}. Thus, improving transformer performance has been in the focus of numerous research and industrial groups. 

Significant attention has been given to optimizing transformers: local and fixed-window attention~\cite{bahdanau2014neural,luong2015effective,shen2018bi,parmar2018image,tay2019simple}, more general structured sparsity~\cite{child2019generating}, learned sparsity~\cite{correia2019adaptively,sukhbaatar2019adaptive,tay2020sparse}, and other algorithmic techniques~\cite{lan2019albert,kitaev2020reformer} improve the performance of transformers. Major hardware efforts, such as Tensor Cores and TPUs~\cite{jouppi2017datacenter} have accelerated tensor operations like matrix-matrix multiplication (MMM), a core transformer operation. Despite this,  \textbf{existing implementations do not efficiently utilize GPUs}. Even optimized implementations such as Megatron~\cite{shoeybi2019megatron} report achieving only 30\% of peak GPU floating point operations per second (flop/s).

We find that \textbf{the key bottleneck when training transformers is data movement}. Improvements in compute performance have reached the point that, due to Amdahl's Law and the acceleration of tensor contractions, training is now memory-bound. Over a third (37\%) of the runtime in a BERT training iteration is spent in memory-bound operators: While tensor contractions account for over 99\% of the arithmetic operations performed, they constitute only 61\% of the runtime. By optimizing these, we show that the overhead of data movement can be reduced by up to 22.91\%. Further, while MMM is highly tuned by BLAS libraries and hardware, we also find that \textbf{existing frameworks use suboptimal data layouts}. Using better layouts enables us to speed up MMM by up to 52\%. Combining these insights requires moving beyond peephole-style optimizations and \textbf{globally optimizing data movement}, as selecting a single layout is insufficient. Overall, we achieve at least $1.30\times$ performance improvements in training over general-purpose deep learning frameworks, and $1.08\times$ over DeepSpeed~\cite{deepspeed}, the state of the art manually-tuned implementation of BERT. For robustly training BERT~\cite{liu2019roberta}, this translates to a savings of over \$85,000 on AWS using PyTorch. For the GPT-3 transformer model~\cite{brown2020language} with a training cost of \$12M~\cite{venture}, our optimizations could save \$3.6M and more than 120 MWh energy. To go beyond this, we develop a recipe for systematically optimizing data movement in DNN training.

Our approach constructs a dataflow graph for the training process, which we use to identify operator dependency patterns and data volume. With this representation, we identify opportunities for data movement reduction to guide optimization. We aim to maximize data reuse using various forms of fusion. Then we select performant data layouts, which is particularly impactful for normalization and tensor contraction operators, where it provides opportunities for vectorization and different tiling strategies. The performance data gathered is then used to find operator configurations that produce an optimized end-to-end implementation.

We evaluate these implementations first for multi-head attention, a core primitive within transformers and one that has significant applications beyond transformers~\cite{bello2019attention,parmar2019stand,cordonnier2019relationship}. We then consider the encoder layer from BERT~\cite{devlin2018bert}, a widely-used transformer architecture. In each case, we compare against existing highly optimized implementations to provide strong baselines. Using this recipe, we are able to demonstrate significant performance improvements in both microbenchmarks and end-to-end training, outperforming PyTorch~\cite{paszke2019pytorch}, TensorFlow+XLA~\cite{tensorflow2015-whitepaper}, cuDNN~\cite{chetlur2014cudnn}, and DeepSpeed~\cite{deepspeed}. While we focus our work on particular transformer models, our approach is generally applicable to other DNN models and architectures.
We summarize our contributions as follows:
\begin{itemize}[itemsep=0.1mm,parsep=0pt]\vspace{-0.5em}
\item We find transformer training to be memory-bound and significantly underperforming on GPUs.
\item We develop a generic recipe for optimizing training using dataflow analyses.
\item Using this recipe, we systematically explore performance of operators and the benefits of different optimizations.
\item We demonstrate significant performance improvements, reducing data movement overheads by up to 22.91\% over existing implementations, and achieving at least $1.08\times$ performance improvements over specialized libraries, and $1.30\times$ over general-purpose frameworks.
\item We make our code publicly available at \scriptsize{\url{https://github.com/spcl/substation}}.
\end{itemize}

\section{Background}
\label{sec:background}
Here we provide a brief overview of our terminology, transformers, and data-centric programming. We assume the reader is generally familiar with training deep neural networks (see~\citet{Goodfellow-et-al-2016} for an overview) and transformers (see Sec.~\ref{subsec:supp-background} for a more complete overview).

\subsection{Transformers}
\label{subsec:bg-transformers-main}

We use the standard data-parallel approach to training, wherein a mini-batch of samples is partitioned among many GPUs. During backpropagation, we distinguish between two stages: computing the gradients with respect to a layer's input ($dX$), and computing the gradients with respect to the layer's parameters ($dW$). Note that the second stage is relevant only for layers with learnable parameters.

\begin{figure}[t]
    \centering
    \includegraphics[width=0.35\textwidth]{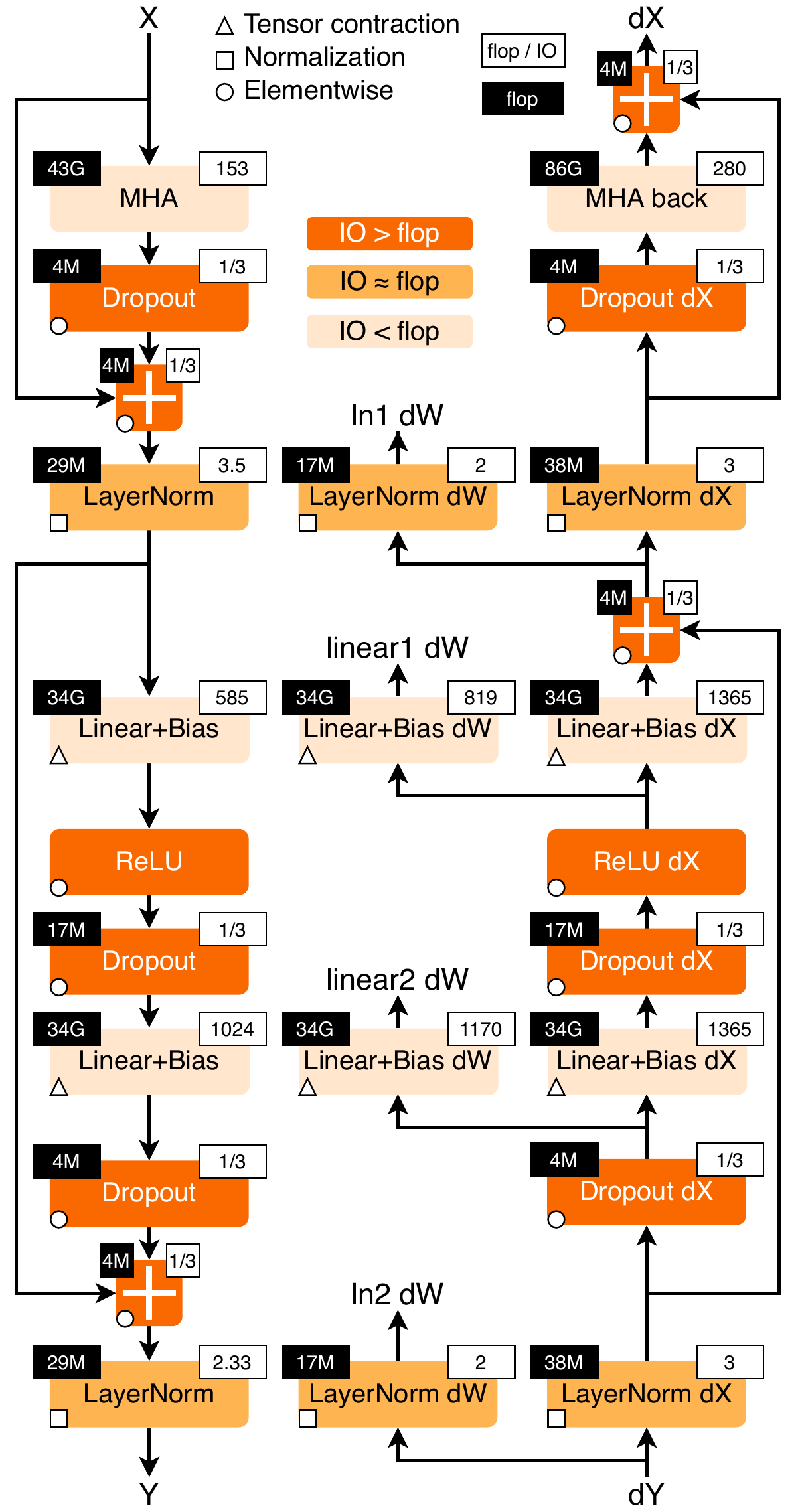}
    \caption{Forward and backpropagation for a BERT-large encoder layer, and the ratio of flop to memory accesses (words) when training on a batch $B=8$ and sequence length $J=K=512$.}
    \vspace{-1em}
    \label{fig:bg-transformer}
\end{figure}

The transformer architecture~\cite{vaswani2017attention} consists of two main components: multi-head attention (MHA) and a feed-forward network.
We provide Python code and an illustration of MHA forward propagation in Fig.~\ref{fig:mha}.
Attention has three input tensors, queries (\texttt{q}), keys (\texttt{k}), and values (\texttt{v}).
The inputs are first multiplied by weight tensors \texttt{wq}, \texttt{wk}, \texttt{wv}, respectively, as a learned input projection (we use Einstein-notation sums, or \texttt{einsums}, for tensor contractions). The query and key tensors are subsequently multiplied together and scaled (stored in \texttt{beta}), followed by a softmax operation.
This is then multiplied with \texttt{vv} to produce the per-head output (\texttt{gamma}).
The outputs of all heads are finally concatenated and linearly projected back to the input dimension size (\texttt{i}).

The respective dataflow graph (Fig.~\ref{fig:mha-sdfg}) immediately exposes coarse- (whole graph) and fine-grained (within rectangular nodes) parallelism, as well as data reuse. As every edge represents exact data movement, their characteristics (access sets and movement volume) can be inspected for bottlenecks and potential solution analysis.

We focus on BERT~\cite{devlin2018bert}.
Fig.~\ref{fig:bg-transformer} illustrates the forward and backward pass for a single BERT encoder layer. The layer essentially consists of MHA followed by a feed-forward neural network (two linear layers with bias and ReLU activations after the first layer). Dropout~\cite{srivastava2014dropout}, layer normalization~\cite{ba2016layer}, and residual connections~\cite{he2016deep} are also used.

\subsection{Data-Centric Programming}
\label{subsec:bg-datamvmt}

As DNN processing is among the most popular compute-intensive applications today, considerable efforts have been made to optimize its core operators \cite{ddlsurvey}. This has driven the field to the point where optimization today is almost exclusively performed beyond the individual operator, either on the whole network~\cite{xla,torchscript} or repeated modules.

Performance optimization on modern architectures consists of mutations to the original code, sometimes algorithmic \cite{chellapilla2006high,mathieu2013fast,lavin2016fast} but mostly relating to hardware-specific mapping of computations and caching schemes. This includes tiling computations for specific memory hierarchies, using specialized units (e.g., Tensor Cores) for bulk-processing of tiles, modifying data layout to enable parallel reductions, hiding memory latency via multiple buffering, pipelining, and using vectorized operations. It is thus apparent that all current optimization techniques revolve around careful tuning of data movement and maximizing data reuse.

The Data-Centric (DaCe) parallel programming framework~\cite{dace} enables performance optimization on heterogeneous architectures by defining a development workflow that enforces a separation between computation and data movement. The core concept powering program transformation is the Stateful Dataflow multiGraph (SDFG), a graph intermediate representation that defines containers and computation as nodes, and data movement as edges. DaCe takes input code written in Python or DSLs, and outputs corresponding SDFGs. Programmers can mutate the dataflow via user-extensible graph-rewriting transformations to change the schedule of operations, the layout of data containers, mapping of data and computation to certain processing elements, or any adaptation to the data movement that does not change the underlying computation.

As opposed to traditional optimizing compilers and deep learning frameworks (e.g., XLA, TorchScript), DaCe promotes a white-box approach for performance optimization. The framework provides an API to programmatically instrument and explore, e.g., layouts and kernel fusion strategies, without modifying the original code. DaCe can map applications to different hardware architectures, including CPUs, GPUs, and FPGAs~\cite{dace}, enabling both whole-program and micro-optimizations of nontrivial applications to state-of-the-art performance~\cite{dace-omen}.

The combination of the separation of the algorithm from the transformed representation and white-box approach for optimization enables us to inspect and optimize the data movement characteristics of Transformer networks. As we shall show in the next sections, this drives a methodical approach to optimizing a complex composition of linear algebraic operations beyond the current state of the art.

\section{Optimizing Transformers}
\label{sec:elements}
We now apply our recipe to optimize data movement in training, using a BERT encoder layer as an example. We focus on a single encoder layer, since these are simply repeated throughout the network, and other components (e.g., embedding layers) are not a significant component of the runtime. In this section, we discuss dataflow and our operator classification. Sections \ref{sec:fusion} and \ref{sec:data-layout} discuss our optimizations and Section~\ref{sec:e2e} presents end-to-end results for transformers.

At a high level, our recipe consists of the following steps:\vspace{-0.5em}
\begin{enumerate}[itemsep=0.1mm,parsep=0pt]
\item Construct a dataflow graph for training and analyze the computation to identify common operator classes.
\item Identify opportunities for data movement reduction within each operator class using data reuse as a guide.
\item Systematically evaluate the performance of operators with respect to data layout to find near-optimal layouts.
\item Find the best configurations to optimize end-to-end performance of the training process.
\end{enumerate}\vspace{-0.5em}

\subsection{Dataflow Analysis}
\label{subsec:ele-dataflow}

We use SDFGs and the DaCe environment to construct and analyze dataflow graphs. Fig.~\ref{fig:bg-transformer} provides a simplified representation of dataflow in a transformer encoder layer. Each node represents an \emph{operator}, which is a particular computation along with its associated input and output, which may vary in size. An operator may be implemented as multiple compute kernels, but is logically one operation for our analysis. To produce an SDFG, all that is required is a simple implementation using NumPy. As the goal of this stage is to understand the dataflow, we do not need to optimize this implementation: It is simply a specification of the computations and data movement.

Using DaCe, we can estimate data access volume and the number of floating point operations (flop) required for each computation. Fig.~\ref{fig:bg-transformer} is annotated with flop and the ratio of flop to data volume, and we provide a precise comparison with PyTorch in Tab.~\ref{tab:flop-analysis}. The key observation is that the ratio of data movement to operations performed varies significantly among operators. In many cases, the runtime of an operator is dominated by data movement, rather than computation, and this should be the target for optimization.


\subsection{Operators in Transformers}
\label{subsec:ele-ops}

\begin{table}
  \caption{Proportions for operator classes in PyTorch.}
  \label{tab:ele-flop-prop}
  \centering
  \small
  \begin{tabular}{lll}
    \toprule
    Operator class & \% flop & \% Runtime \\
    \midrule
    \tenscons{} Tensor contraction & 99.80 & 61.0 \\
    \statnorm{} Stat. normalization & 0.17 & 25.5 \\
    \elewise{} Element-wise & 0.03 & 13.5 \\
    \bottomrule
  \end{tabular}
  \vspace{-1em}
\end{table}

With this analysis, we can now identify high-level patterns that allow us to classify operators. We base our classification both on the data movement to operation ratio and the structure of the computations. This classification is useful as it allows us to consider optimizations at a more general level, as opposed to working on each operator (or kernel) individually. For transformers, we find three classes: tensor contractions, statistical normalizations, and element-wise operations. The border of each operator in Fig.~\ref{fig:bg-transformer} indicates its class and Tab.~\ref{tab:ele-flop-prop} gives the proportion of flop and runtime for a BERT encoder layer for each class.

\textbf{Tensor Contractions \tenscons{}} These are matrix-matrix multiplications (MMMs), batched MMMs, and in principle could include arbitrary tensor contractions. We consider only MMMs and batched MMMs for simplicity, as these are efficiently supported by cuBLAS. In transformers, these are linear layers and components of MHA. These operations are the most compute-intensive part of training a transformer. For good performance, data layout and algorithm selection (e.g., tiling strategy) are critical.

\textbf{Statistical Normalizations \statnorm{}} These are operators such as softmax and layer normalization. These are less compute-intensive than tensor contractions, and involve one or more reduction operation, the result of which is then applied via a map. This compute pattern means that data layout and vectorization is important for operator performance.

\textbf{Element-wise Operators \elewise{}} These are the remaining operators: biases, dropout, activations, and residual connections. These are the least compute-intensive operations.

\subsection{Memory Usage Efficiency (MUE)}

The MUE metric~\cite{fuhrer2018near} provides a measure of the memory efficiency of an operation, both with respect to its implementation and achieved memory performance. This provides another method for understanding performance beyond flop/s that is particularly relevant for applications that are bottlenecked by data movement. MUE evaluates the efficiency of a particular \emph{implementation} by comparing the amount of memory moved ($D$) to the theoretical I/O lower bound~\cite{jia1981complexity} ($Q$) and the ratio of achieved ($B$) to peak ($\hat{B}$) memory bandwidth: $\mue = \sfrac{Q}{D} \cdot \sfrac{B}{\hat{B}} \cdot 100$.
If an implementation both performs the minimum number of operations and fully utilizes the memory bandwidth, it achieves $\mue = 100$. This can be thought of as similar to the percent of peak memory bandwidth.

\subsection{Evaluation Details}
\label{subsec:ele-eval-details}

We use the Lassen supercomputer~\cite{lassen}, where each node has four 16 GB Nvidia V100 GPUs with NVLINK2. For comparison, we use cuDNN 8.0.4, PyTorch 1.6.0 (PT), TensorFlow 2.1 from IBM PowerAI with XLA, and a recent development version of DeepSpeed (DS). See Sec.~\ref{subsec:supp-eval-details} for more details. Our running example is training BERT-large. We use a mini-batch size of $B=8$, sequence length $L=512$, embedding size $N$=$1024$, $H$=$16$ heads, and a projection size $P=64$.


\section{Fusion}
\label{sec:fusion}

\begin{figure}
    \centering
    \includegraphics[width=0.38\textwidth]{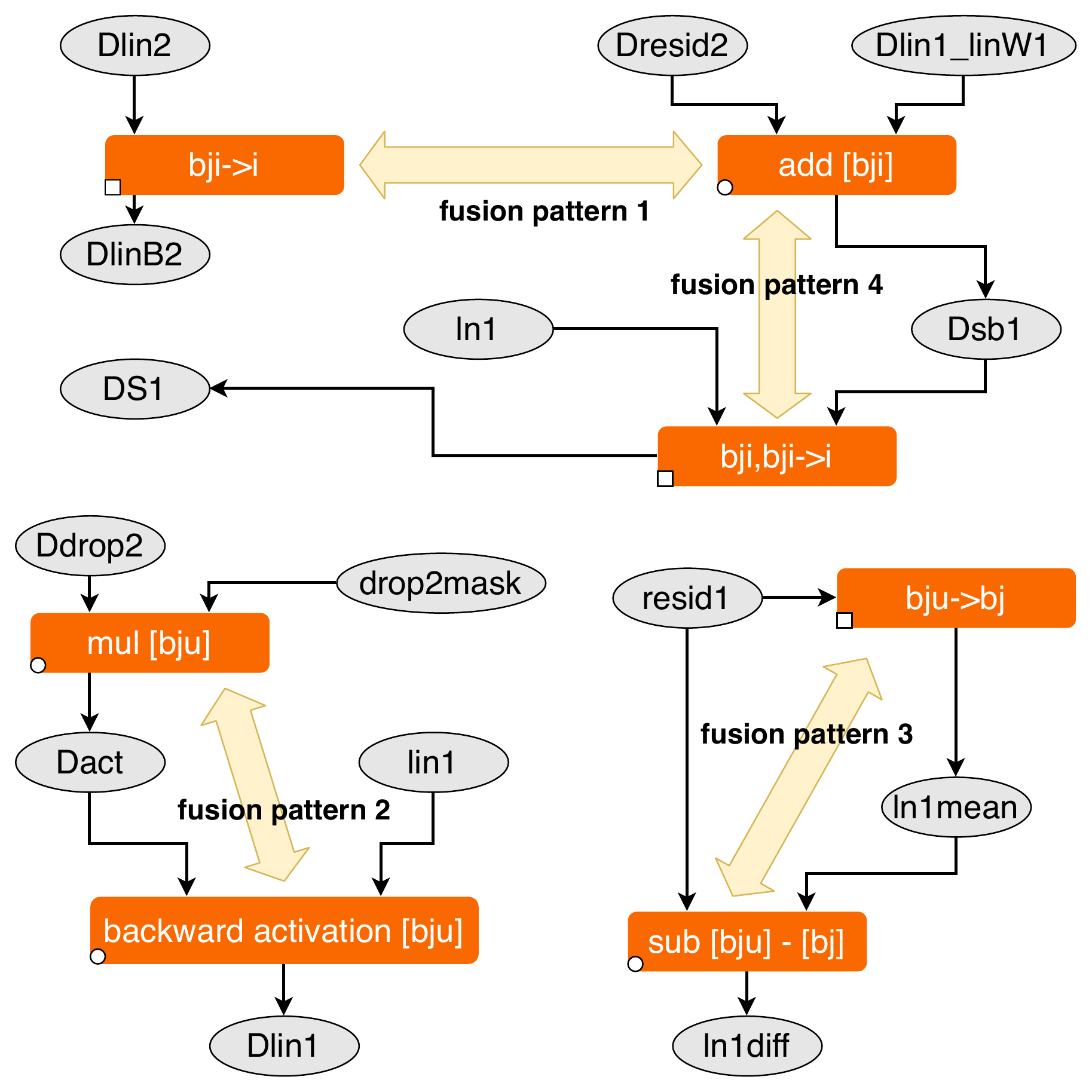}
    \vspace{-0.5em}
    \caption{Examples of operator fusion.}
    \vspace{-1em}
    \label{fig:fusion-examples}
\end{figure}

A significant portion of the runtime in existing transformer implementations is in statistical normalization and element-wise operators (Tab.~\ref{tab:ele-flop-prop}). Thus, fusion is a major opportunity for promoting data reuse, as when operators cover identical iteration spaces, global memory writes and subsequent reads between them can be removed.



We develop a set of fusion rules applicable to any application with operators similar to the three here. The process consists of two steps: detecting which operations can be fused and then deciding whether it is beneficial to fuse them.

To discover fusion opportunities, we analyze iteration spaces. Every operator has independent dimensions. Statistical normalization operators also have reduction dimensions; tensor contractions also have reduction dimensions and special independent dimensions for the two input tensors.


The type of \textit{iteration space implementation} determines which tools are used to make them. Independent dimensions can be implemented using GPU block or thread parallelism, or "for" loops within a single thread. Reduction dimensions use these techniques but also specify how the reduction is to be performed: accumulating to the same memory location ("for" loops), or as grid-, block-, or warp-reductions.


Two operators can be fused if their \textit{iteration space implementations} are compatible: They are either the same or the only difference is that one operator performs a reduction. The order and size of dimensions and the implementation for each must match. If the first operator produces output the second uses as input, partial fusion can be done: The outermost independent dimensions can be shared, but the innermost iteration spaces are put in sequential order inside.

When a fusion opportunity is detected, we take it in two cases: First, if we can perform fewer kernel launches by merging iteration spaces. Second, if we achieve less data movement by avoiding loads and stores between operators. Theoretically, the first case could increase memory pressure, but we observe it provides benefits in practice.

We attempt to fuse maximally. There are four structural patterns (Fig.~\ref{fig:fusion-examples}) in the dataflow graph for the encoder layer when fusion rules above are applied to a pair of non-tensor contraction operators. Using the SDFG, we fuse two adjacent operators whenever we detect these patterns and continue until we cannot fuse further. This means we fuse until either a reduction dimension or iteration space changes. As a further constraint, we only fuse simple element-wise scaling operations into tensor contraction operators.

Each fused operator is implemented as a CUDA kernel specialized for a specific data layout. Due to space limitations, we detail our implementation and fused kernels in Sec.~\ref{subsec:fusion-impl}.

\subsection{Results}
\label{subsec:fusion-results}
 

Tab.~\ref{tab:flop-analysis} presents our comprehensive results, including operator fusion. In this, we show a detailed breakdown of the required and observed flop, data movement, runtime, and MUE for each operator within the encoder layer, for both PyTorch and our implementation, with our fused operators marked. We can easily observe that while the vast majority of flop are in tensor contractions, much of the runtime is in statistical normalization and element-wise operators (see also Tab.~\ref{tab:ele-flop-prop}). These operators are indeed memory-bound.

In forward propagation, every fused operator outperforms PyTorch's. In backpropagation, this trend generally holds, but \textsc{ebsb} and \textsc{baob} are slower due to our configuration selection algorithm choosing a suboptimal layout for some operators to optimize the overall performance (see Sec.~\ref{sec:e2e}). 

By studying the MUE and flop/s, we can reason about the bottlenecks behind each operator. For the fused operators, we see that  high MUE rates are often achieved. In fact, the MUE from Tab.~\ref{tab:flop-analysis} and the theoretical flop/IO ratio from Fig.~\ref{fig:bg-transformer} are highly correlated across operators. We say that a kernel is memory bound if its MUE is larger than the achieved peak flop/s, and compute bound otherwise. This insight aids in analyzing the bottlenecks of general DNNs and automated tuning of operators, prior to measuring their performance. We note that for our operators, which involve multiple tensors of different shapes, 100\% MUE is potentially unattainable as achieving peak memory bandwidth requires a highly regular access pattern into DRAM.

As for the tensor contraction results, we see that the attained MUE is consistently under 50\%. This is acceptable, since the underlying matrix multiplications are generally compute-bound. However, we note that some cases, such as $QK^T$, are both low in flop/s \textit{and} MUE. A more in-depth look into the dimensions of the contraction reveals that the dimensions are small, which then indicates that the tensor cores are underutilized. This may result from insufficient scheduled threads, or low memory throughput to compute ratio. We thus try to increase hardware utilization by fusing additional operators into the contractions next.

\textbf{Additional Fusion Approaches} We considered two additional fusion approaches, fusing operators into tensor contractions and algebraic fusion, which we discuss in Sec.~\ref{subsec:supp-fusion} due to limited space.
We find that fusing into tensor contractions is not profitable, but algebraic fusion to combine input projection operations in MHA is.

\section{Data Layout}
\label{sec:data-layout}
We now consider data layout selection, which enables efficient access patterns, vectorization, and tiling strategies for tensor contraction and statistical normalization operators. To study this systematically, for each operator, including fused operators produced in the prior step, we benchmark every feasible data layout, as well as varying other parameters depending on the specific operator. The best parameterization of an operator is highly dependent on the GPU model and tensor sizes, and may not be obvious a priori; hence it is important to consider this empirically.

Because there are a myriad of potential configurations for each operator, we summarize the distribution of runtimes over all configurations using violin plots. The width of the violin represents the relative number of configurations with the same runtime. This allows us to see not only the best runtime but sensitivity of operators to layouts, an important factor for global layout optimization in Step 4 of our recipe.

\subsection{Tensor Contractions}
\label{subsec:dl-tc}

Using the Einsum notation for tensor contractions, we consider all equivalent permutations of the summation string. The einsum is then mapped to a single cuBLAS MMM or batched MMM call. While this notation allows us to express arbitrary tensor contractions, as cuBLAS does not support all configurations, we limit ourselves to these two types.

In addition, we consider every possible cuBLAS algorithm for each layout, as we have observed that the heuristic default selection is not always optimal. We use the \texttt{cublasGemmEx} API to manually select algorithms. We use both regular and Tensor Core operations, and perform all accumulations at single-precision, in line with best practices for mixed-precision training~\cite{micikevicius2018mixed}.

\begin{figure*}
    \centering
    \includegraphics[width=\textwidth]{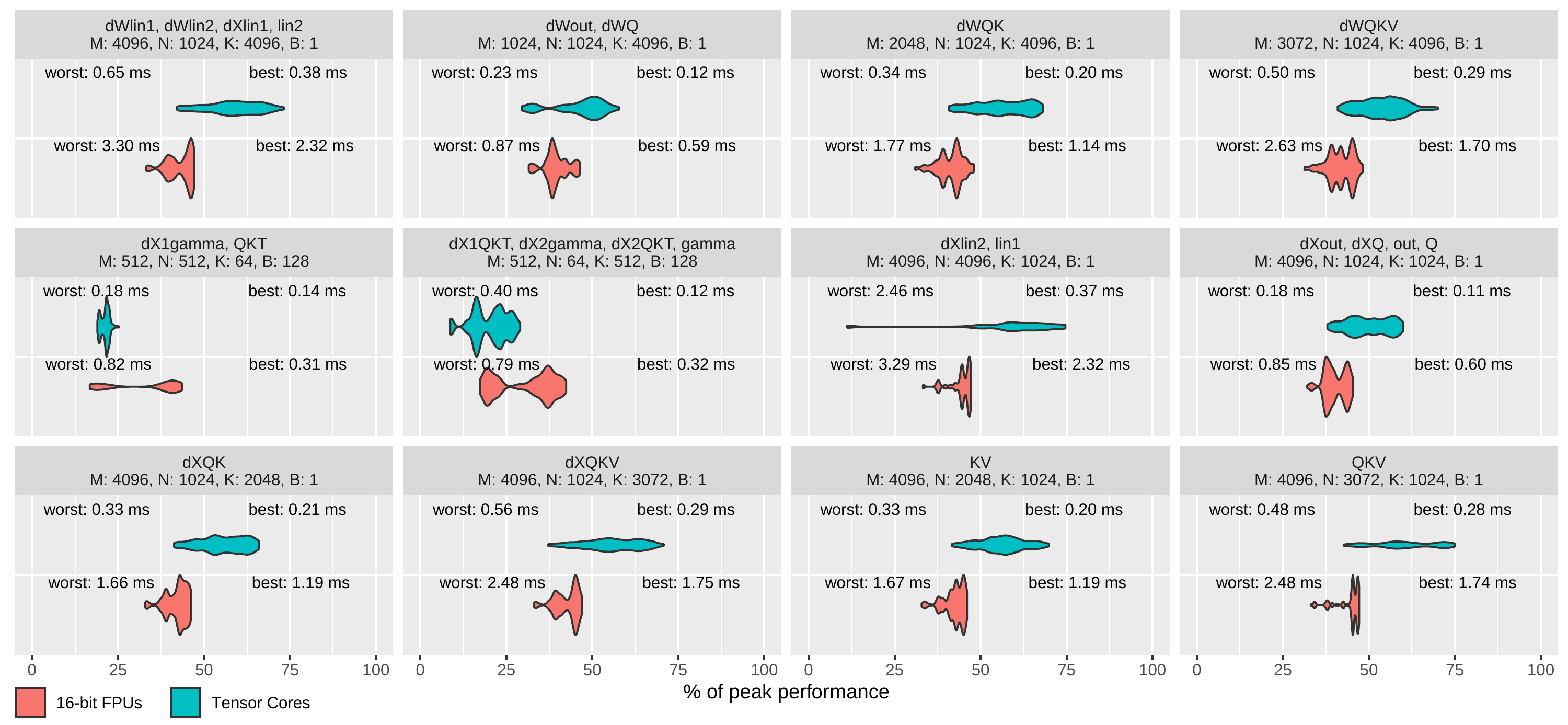}
    \vspace{-2em}
    \caption{Tensor contraction performance. Tensor Core peak: 125 Tflop/s; FP16 peak: 31.4 Tflop/s.}
    \label{fig:dl-tc-violins}
\end{figure*}

Fig.~\ref{fig:dl-tc-violins} presents performance distributions over all data layouts for every tensor contraction in the encoder layer training, including algebraic fusion variants. Each plot is for different tensor sizes and shows the performance with and without Tensor Cores. Since input matrices for cuBLAS MMMs can be easily swapped, results for both orders are merged into the plot and labeled with $M > N$.
Interestingly, in several cases (where $N$ or $K$ is 64) the performance is quite close to the regular floating point units, due to a failure to saturate the tensor cores. Among the tensor core results, we can typically see there are several modes in the performance distributions; these correspond to particularly important axes for data layout. Indeed, for many configurations, one of these is near to or contains the best-performing configuration, indicating that many slightly different data layouts could be used with little impact on performance depending on the needs of our global optimization pass. However, this does not mean that \emph{any} data layout is acceptable; in every case, the majority of the layouts do not perform well, illustrating the importance of careful tuning.

We also investigated how the default cuBLAS algorithm compares to the best-performing configuration. On half precision, we found that the algorithm chosen by cuBLAS's heuristic was up to 14.24\% worse than the best algorithm (in $dX1\,QK^T$). We also investigated the performance at single-precision and found similar results with up to 7.18\% worse performance. This demonstrates the importance of tuning for particular hardware and workload.

\subsection{Fused Operators}
\label{subsec:dl-fused}

For our fused operators, we consider all combinations of input and output layout permutations. This enables us to include transposes of the output data as part of the operator, should the next operator perform better in a different layout than the input. The data layout is especially critical for statistical normalization operators, where the appropriate data layout can enable vectorization opportunities, especially vectorized loads and stores for more efficient memory access. We therefore also consider vectorization dimensions, the mapping of dimensions to GPU warps, etc. Our implementation takes advantage of these layouts when feasible.

\begin{figure}
    \centering
    \includegraphics[width=0.45\textwidth]{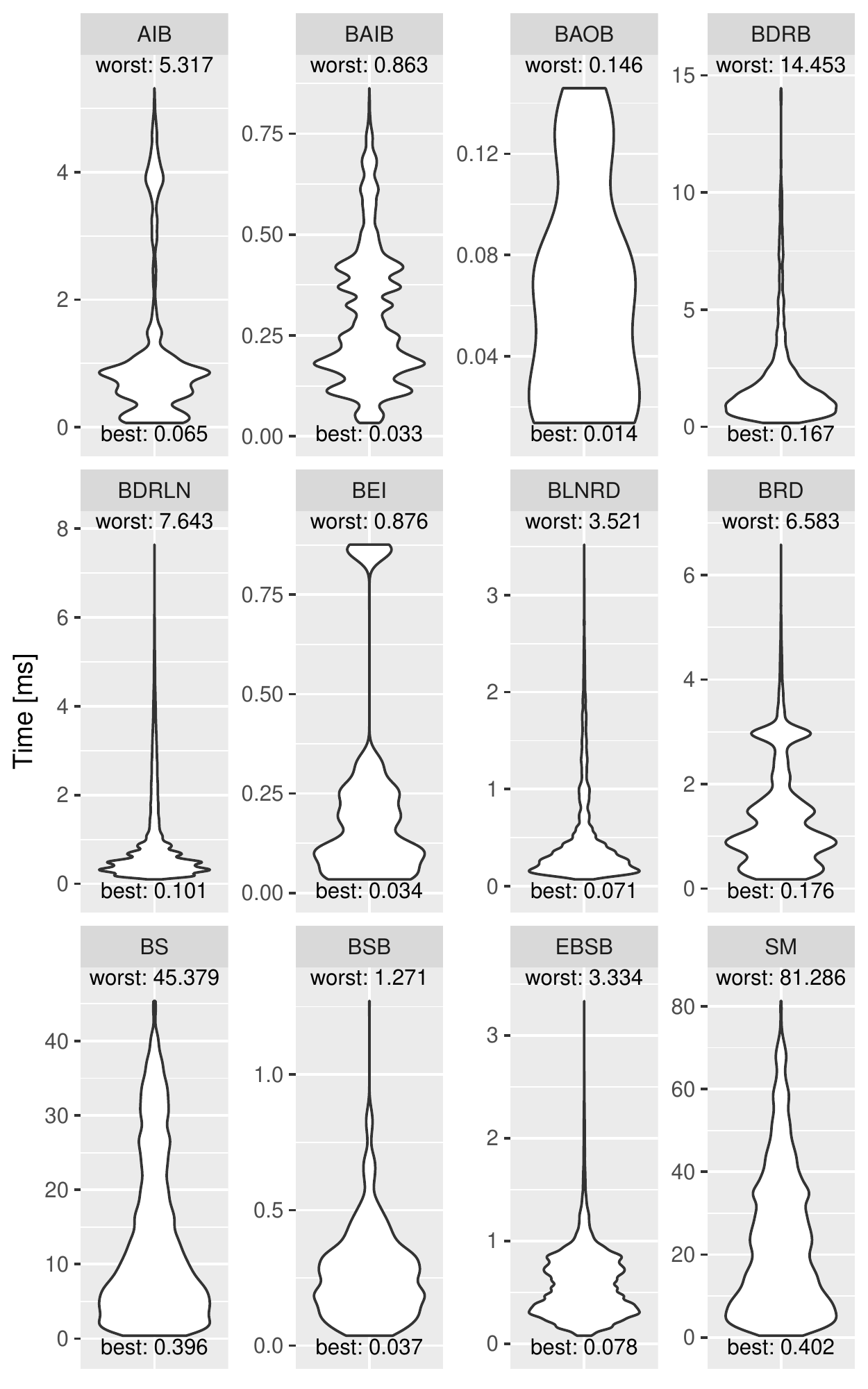}
    \caption{Performance of fused kernels for element-wise and statistical normalization operators.}
    \label{fig:dl-fused-layout-violins}
    \vspace{-1em}
\end{figure}

Fig.~\ref{fig:dl-fused-layout-violins} presents the runtime distribution for all configurations of our fused operators (note that some are used twice; see Sec.~\ref{subsec:fusion-impl} for details of each operator). The distributions here are qualitatively similar to those in Fig.~\ref{fig:dl-tc-violins}, except these have much longer tails: A bad configuration is, relatively, much worse, potentially by orders of magnitude.

All kernels support changing the layouts of tensors they use. This change is done via template parameterization, so the compiler can generate efficient code. All kernels support selecting different vectorization dimensions. The \textsc{brd} and \textsc{bei} kernels can select the dimension used for CUDA thread distribution; \textsc{bsb}, \textsc{ebsb}, and \textsc{bdrb} can select the warp reduction dimension, as they reduce over two dimensions.

The most noticeable performance improvement is made by layouts that enable vectorized memory access, showing that the main performance limitation is the amount of moved data.
The second notable category are layouts with the same reduce and vector dimensions. Joining these dimensions decreases the number of registers required to store partially reduced values from the vector size (eight at FP16) to one.

We can expect to get good performance restricting ourselves to configurations in the two groups described above. Usually, the best layout discovered follows them. For example, the \textsc{sm} kernel has the same warp and reduction dimensions, and these dimensions are the last and sequential ones for involved arrays. However, this intuition is not always correct. 
From the results in Fig.~\ref{fig:dl-fused-layout-violins}, we find that there are configurations that both satisfy these intuitive rules yet are very slow. For example, the best configuration of \textsc{aib} takes 65 $\mu$s, and the worst "intuitively good" configuration takes 771 $\mu$s.

Configurations discovered through exhaustive search can enhance our intuitive understanding of what is required for good performance. For example, the \textsc{brd} kernel uses four 3D tensors, which can use six possible layouts. Intuitively, we want to place the vectorized dimension last to make it sequential for all tensors. Surprisingly, however, the best configuration has only two tensors vectorized. With this information, intuition can be refined: the likely factor that limits vectorization over all tensors is excessive register usage. However, unlike the exhaustive search, intuition does not help to identify \emph{which} tensors to vectorize.

\section{End-to-End Optimization}
\label{sec:e2e}
The final step is to assemble fused components and select data layouts for each operator to yield a complete implementation. This is the culmination of the prior steps performing dataflow analysis, fusion, and layout selection. From these, we have performance data for all data layouts and algebraic fusion strategies. One cannot simply pick a single layout a priori, as the benefit of running two operators in different layouts may outweigh the overhead of transposing data.
Our assembled implementation is structured using the SDFGs produced in Step 1. We integrate it into PyTorch~\cite{paszke2019pytorch} via its C++ operator API.

\subsection{Configuration Selection}
\label{subsec:e2e-config-selection}

\begin{figure}[t]
    \centering
    \includegraphics[width=0.4\textwidth]{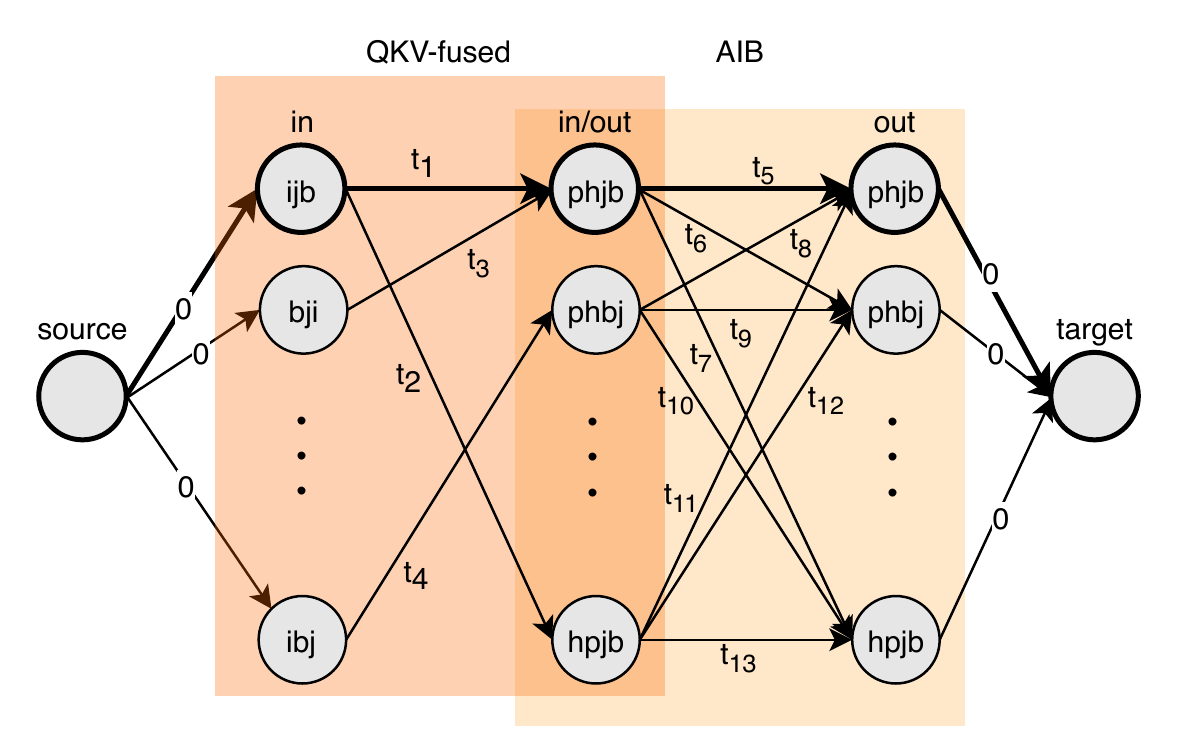}
    \vspace{-1em}
    \caption{Example configuration selection graph for SSSP.}
    \vspace{-1.5em}
    \label{fig:confsel}
\end{figure}

We develop an automatic configuration selection algorithm to globally optimize our implementation using the performance data. We construct a directed graph (Fig.~\ref{fig:confsel}) based on the dataflow graph of the operators. Beginning from the input data and proceeding in a topological order, we add a node to the graph for each input and output data layout of the operator. An edge is added from the input to the output layout, weighted with the minimum runtime of any configuration with that layout. Determining this simply requires a linear search over the matching performance data. This allows us to select both the best data layout and any other operator knobs (e.g., vectorization dimension). To minimize the size of the graph, we only add a configuration from an operator when it has at least one input and output edge. We then run a single-source shortest-path (SSSP) algorithm from the input to the output in the graph; the resulting path gives our final configuration. Because this graph is a DAG and the number of feasible input/output configurations for each operator is small, SSSP takes linear time asymptotically and seconds for BERT. The path is saved to a configuration file which is used to automatically define tensor layouts at the start of training.

To simplify the implementation, we omit dataflow connections between forward and backpropagation operators. This assumption could be relaxed in a future version of this algorithm. Although this means we are not guaranteed to find a globally optimal data layout, the runtime of our configuration is nevertheless within 6\% of an ideal (incorrect) layout configuration that ignores data layout constraints.

\subsection{Multi-head Attention}
\label{subsec:e2e-mha}

We first analyze the performance of multi-head self-attention. While it is a key primitive in BERT, MHA is also used outside of transformers, so understanding its performance in isolation can inform other models too. Tab.~\ref{tab:eval-mha-time} compares our globally-optimized implementation with PyTorch, TensorFlow+XLA, and cuDNN. cuDNN's MHA implementation (\texttt{cudnnMultiHeadAttnForward} and related) supports six data layouts; we report the fastest.

cuDNN's performance is significantly worse than the others; as it is a black box, our ability to understand it is limited. However, profiling shows that its implementation launches very large numbers of softmax kernels, which dominate the runtime, indicating additional fusion would be profitable. TensorFlow+XLA finds several fusion opportunities for softmax. However, its implementation does not perform algebraic fusion for the queries, keys, and values, and it uses subpar data layouts for tensor contractions.

Our performance results in Tab.~\ref{tab:flop-analysis} illustrate the source of our performance advantage over PyTorch: Our data layout and algorithm selection results in faster tensor contractions overall. This is despite the Gamma stage actually being slower than PyTorch's: Sometimes locally suboptimal layouts need to be selected to improve performance globally.

\begin{table}[t]
  \caption{Multi-head attention performance for BERT.}
  \label{tab:eval-mha-time}
  \centering
  \small
  \begin{tabular}{lllll}
    \toprule
    & TF+XLA & PT & cuDNN & Ours \\
    \midrule
    Forward (ms) & 1.60 & 1.90 & 3.86 & \textbf{1.25} \\
    Backward (ms) & 2.25 & 2.77 & 680 & \textbf{1.86} \\
    \bottomrule
  \end{tabular}
\end{table}

\subsection{End-to-End Performance}
\label{subsec:e2e-e2e}





We present overall performance results for the encoder layer in Tab.~\ref{tab:eval-encoder-time}. For forward and backpropagation combined, we are $1.30\times$ faster than PyTorch and $1.20\times$ faster than TensorFlow+XLA, including  framework overheads. At a high level, this is because we perform a superset of the optimizations used by \emph{both} frameworks, and globally combine them to get all the advantages while minimizing drawbacks. As a general guideline, we use flop and MUE rates as proxies for which operators require the most attention and their corresponding bottlenecks. This ensures a guided optimization rather than tuning all operators aggressively.

We also include performance results from DeepSpeed, which we are $1.08\times$ faster than. This is despite DeepSpeed being a manually-optimized library tuned specifically for BERT on V100s. Note also that DeepSpeed modifies some operations, e.g., to be reversible or to exploit output sparsity, and so is not always strictly comparable to the other implementations. This also provides it opportunities for optimization that we do not consider.

The total data movement reduction we attain is $\sim$22.91\% over the standard implementation. We obtain this information from Tab.~\ref{tab:flop-analysis}, where for each fused kernel we omit the interim outputs and inputs that are not part of the overall I/O that the fused kernels perform. TensorFlow+XLA's automatic kernel fusion reduces data movement similarly to ours. However, the data layouts used for tensor contractions are not optimal, and its BERT encoder implementation does not use algebraic fusion in MHA. PyTorch's data layouts enable faster tensor contractions and it implements the algebraic fusion, but it has higher overheads for other operators.

Our fusion pass finds all the opportunities that TF+XLA does, plus several additional ones; for example, we implement layernorm as a single fused kernel. Our data layout selection picks better layouts than PyTorch in nearly every individual instance; when it does not, this is because the layout change enables greater performance gains downstream. In Tab.~\ref{tab:flop-analysis}, we also see that PyTorch performs more flop than predicted. Some of this is due to padding in cuBLAS operations, and generic methods performing excess operations. However, we also discovered that some cuBLAS GEMM algorithms, including ones called by PyTorch, incorrectly perform twice as many FLOPs as necessary; our recipe avoids these cases automatically.

We also considered another configuration for training BERT, where we change the batch size to $B=96$ and the sequence length to $L=128$, and retuned our layout selection. In this case, forward and backpropagation for a single encoder layer takes 18.43 ms in PyTorch, 16.19 ms in DeepSpeed, and 15.42 ms in our implementation. We outperform both PyTorch and DeepSpeed in this case (even with its additional optimizations). We believe that with improvements to our layout selection algorithm, the performance of our implementation will improve further.

\begin{figure}[t]
  \centering
  \includegraphics[height=1in]{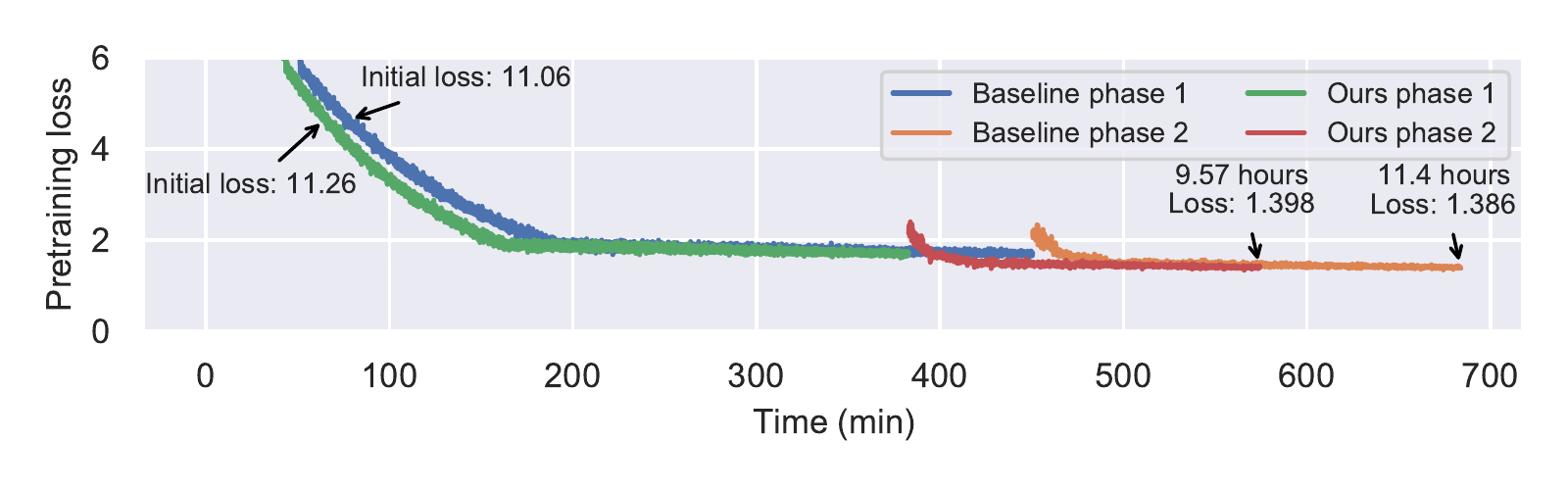}
  \caption{Pretraining curve for BERT-large.}
  \label{fig:e2e-pretraining}
\end{figure}

Finally, we performed end-to-end training of BERT at scale. We observe an overall speedup of $1.19\times$, despite additional training overheads. We show pretraining curves in Fig.~\ref{fig:e2e-pretraining} and present full details of the experiment in Sec.~\ref{subsec:supp-e2e}.



Beyond BERT, other transformers have very similar layers, such as decoder layers in GPT-2/3. With very few changes, our recipe and implementations are directly applicable to these. Our implementation can also be extended to support a full training pipeline by stacking our optimized layers.




\begin{table}[t]
  \caption{Full BERT encoder layer performance.}
  \label{tab:eval-encoder-time}
  \centering
  \small
  \begin{tabular}{lllll}
    \toprule
    & TF+XLA & PT & DS & Ours \\
    \midrule
    Forward (ms) & 3.2 & 3.45 & 2.8 & \textbf{2.63} \\
    Backward (ms) & 5.2 & 5.69 & 4.8 & \textbf{4.38} \\
    \bottomrule
  \end{tabular}
  \vspace{-1em}
\end{table}

\section{Related Work}
\label{sec:related}
We provide a brief overview of related work here, and a more comprehensive discussion in Sec.~\ref{subsec:supp-related}.

Most directly relevant are other approaches specifically to accelerate transformer training. Distributed-memory techniques, such as ZeRO~\cite{rajbhandari2019zero}, Megatron~\cite{shoeybi2019megatron}, and Mesh-TensorFlow~\cite{shazeer2018mesh} scale training to many GPUs to accelerate it. Mesh-TensorFlow also presents a classification of operators similar to ours. Large batches have also been used to accelerate training via LAMB~\cite{you2019large} or NVLAMB~\cite{nvlamb}. None of these directly address the performance of a single GPU as done in this paper. DeepSpeed~\cite{deepspeed}, which we compare with in Section~\ref{subsec:e2e-e2e}, is closest to our work, but performs all optimizations and layout selections manually.

Many frameworks provide optimizations that can also be applied to transformers, such as XLA or TVM~\cite{chen2018tvm}. None of these works provide all the optimizations or the systematic study of data movement and its impact on performance provided here.
Beyond deep learning, data movement reduction is a core component of high-level optimization~\cite{tpds-locality} and polyhedral optimizations (e.g., \citet{polly}).
Other white-box approaches for optimization exist (e.g., Halide~\cite{ragan2013halide} and MLIR~\cite{lattner2020mlir}).
The data-centric optimization approach proposed here using dataflow graphs allows us to perform and tune complex data layout and fusion transformations that span granularity levels, surpassing the optimization capabilities of the aforementioned tools and achieving state-of-the-art performance.

\section{Discussion}
\label{sec:beyond}
The recipe we propose in this paper can be directly adopted in other DNN architectures. Additional transformer networks, such as Megatron-LM~\cite{shoeybi2019megatron} and GPT-3~\cite{brown2020language}, only differ by dimensions and minor aspects in the encoder and decoder blocks (e.g., dropout position, biases). Once a data-centric graph is constructed from them, the recipe remains unchanged.

\subsection{Beyond Transformers}

The classification of operators into three groups covers a wide variety of operators that span beyond transformers.

Large tensors and their contraction are ubiquitous in modern DNNs. 
For MLPs and recurrent neural networks (RNNs), there is little change, as the core operator types are essentially the same.
Convolutions, pooling, and other local spatial operators can be treated similarly to tensor contractions, owing to their arithmetic intensity properties and abundance of optimized implementations. Therefore, the same considerations we take here are just as critical in CNNs. However, as opposed to contractions (see Section \ref{subsec:fusion-fuse-tc}), further fusion is typically considered for convolutions.

Statistical normalization also takes different forms in DNNs. This includes a variety of reductions, as well as Instance, Group, and Batch Normalization, where the latter constitutes the second largest computation in ResNets after convolutions. When varying data layouts, these operators share properties (normalizing a dimension) and are optimized in exactly the same way.
Lastly, element-wise operators always exist in DNNs and benefit from the same fusion and bulk data movement optimizations as we perform here.
For graph neural networks~\cite{bronstein2017geometric}, capsule neural networks~\cite{sabour2017dynamic}, and other emerging architectures, the operators change more significantly, but the basic procedure applies. 

Due to the prohibitively large search space of configurations in transformers, writing manually-optimized kernels becomes infeasible. Instead, each of our data-centric implementations chooses an optimization scheme (e.g., tiling, vectorization, warp-aggregated reductions) automatically, according to the input data layout and the operator type.
Combined with automated configuration selection (Section~\ref{subsec:e2e-config-selection}), we rely only on the dataflow graph structure to choose the best \textit{feasible} data layout configuration.



\subsection{Hardware Implications}

The implications of data movement reduction extend beyond software. Given that the best performance for different operators is achieved with different data layouts, there would be significant benefits if future machine learning accelerators included built-in support for fast data layout changes.
We confirm this in our MUE results (Tab.~\ref{tab:flop-analysis}), as even the most compute-intensive tensor contractions are bounded by the hardware's capability of transferring data to Tensor Cores.

Hardware trends indicate a similar situation. New architectures are moving towards reducing data format conversion (e.g., TF32~\cite{tf32}), increased on-chip memory and low-latency interconnects~\cite{graphcore}, and coarse-grained spatial hardware~\cite{cerebras}. For the latter two, the recipe and analysis provided here is crucial to maintain high utilization in pipelined DNN execution.
More generally, we expect our recipe to be applicable to any load-store architecture (e.g., TPUs~\cite{jouppi2017datacenter}). The \% of peak and MUE are fundamental quantities and analyzing them will allow one to study bottlenecks and optimize appropriately.

\section{Conclusions}
\label{sec:conclusions}
Despite the importance of transformer neural networks, training them is memory-bound and underutilizes GPUs. Using our recipe for data movement analysis, we identified bottlenecks and optimizations, yielding improved implementations that outperform the already highly tuned state-of-the-art. As training transformers is already a major  workload that will only grow larger, our improvements offer significant real-world impacts for both research and industry.

Our approach is applicable more broadly to deep learning; many DNNs easily fit within our operator classification. This is especially important for guiding the optimization of emerging model architectures, which do not benefit from existing acceleration libraries. Our results also highlight the importance of considering data movement at every level of the training stack---from the application down to hardware.

\section*{Acknowledgments}
This project received funding from the European Research Council (ERC) under the European Union's Horizon 2020 programme (grant agreements DAPP, No. 678880, and EPiGRAM-HS, No. 801039). N.D. is supported by the ETH Postdoctoral Fellowship. T.B.N. is supported by the Swiss National Science Foundation (Ambizione Project \#185778). Experiments were performed at the Livermore Computing facility.

\bibliography{paper}
\bibliographystyle{mlsys2021}

%

\clearpage
\appendix
\counterwithin{figure}{section}
\counterwithin{table}{section}
\section{Supplementary Material}
\subsection{Additional Background}
\label{subsec:supp-background}

We use the standard data-parallel approach to training, wherein a mini-batch of samples is partitioned among many GPUs. During backpropagation, we distinguish between two stages: computing the gradients with respect to a layer's input ($dX$), and computing the gradients with respect to the layer's parameters ($dW$). Note that the second stage is relevant only for layers with learnable parameters.



\subsubsection{Transformers}
\label{subsec:bg-transformers}

The transformer architecture~\cite{vaswani2017attention}, originally developed for machine translation, is a neural network architecture for \emph{sequence transduction}, or transforming an input sequence into an output sequence. Transformers build upon a long sequence of work within natural language processing, most relevantly beginning with word embeddings~\cite{mikolov2013linguistic,mikolov2013efficient,mikolov2013distributed}, neural machine translation~\cite{kalchbrenner2013recurrent,cho2014learning}, and sequence-to-sequence learning~\cite{sutskever2014sequence}. A key component is \emph{attention}~\cite{bahdanau2014neural,luong2015effective}, which enables a model to learn to focus on particular parts of a sequence.

The transformer makes two key contributions. First, it generalizes attention to multi-head attention, which we discuss below. Second, the transformer neglects recurrent or convolutional mechanisms for processing sequences, and relies entirely on attention. Critically, this enables significantly more parallelism during training, as the model can process every element of a sequence simultaneously, instead of having a serial dependence on the prior element.

\subsubsection{Multi-head Attention}
Multi-head attention (MHA) generalizes attention, and uses \texttt{h} attention ``heads'' in parallel to attend to different learned projections of a sequence. We provide Python code and an illustration of MHA forward propagation in Fig.~\ref{fig:mha}.

Each attention head is an instance of \emph{scaled dot-product attention}, with input tensors: queries (\texttt{q}), keys (\texttt{k}), and values (\texttt{v}). Conceptually, attention finds values corresponding to the keys closest to the input queries. Heads are augmented with learned linear layers that project their inputs into a lower-dimensional space.
The three inputs are first multiplied by weight tensors \texttt{wq}, \texttt{wk}, \texttt{wv}, respectively, as a learned input projection (we use Einstein-notation sums, or \texttt{einsums}, for tensor contractions). The query and key tensors are subsequently multiplied together and scaled (stored in \texttt{beta}), followed by applying the softmax operation in order to weight and select the most important results.
This is then multiplied with \texttt{vv} to produce the per-head output (\texttt{gamma}).
The outputs of all heads are finally concatenated and linearly projected back to the input dimension size (\texttt{i}).

The respective dataflow graph (Fig.~\ref{fig:mha-sdfg}) immediately exposes coarse- (whole graph) and fine-grained (within rectangular nodes) parallelism, as well as data reuse. As every edge represents exact data movement, their characteristics (access sets and movement volume) can be inspected for guided bottlenecks and potential solution analysis.

There are three broad classes of MHA, distinguished by their inputs. General attention uses distinct tensors as the queries, keys, and values. Encoder/decoder attention uses the same tensor for both keys and values.
Self-attention uses the same tensor for all three inputs. MHA may also have a masking step, which is used during training to prevent a model from ``seeing the future'' and using information from a later part of a sequence.

\subsubsection{Transformer Architectures}

BERT~\cite{devlin2018bert} is a widely-used transformer for NLP tasks. Fig.~\ref{fig:bg-transformer} illustrates the forward and backward pass for a single BERT encoder layer. The layer essentially consists of MHA (as self-attention) followed by a feed-forward neural network (two linear layers with bias and ReLU activations after the first layer). Dropout~\cite{srivastava2014dropout}, layer normalization~\cite{ba2016layer}, and residual connections~\cite{he2016deep} are also used.

Transformers also incorporate several other layers that we will not discuss in detail: embedding layers for input sequences and various output layers, depending on the task. Other transformer architectures, such as the original Transformer and GPT-2/3~\cite{radford2019language,brown2020language} have very similar architectures.

\subsection{Fusion Implementation}
\label{subsec:fusion-impl}

We implement each fused operator as a single custom CUDA kernel and specialize it for a specific data layout using templates to maximize opportunities for compiler optimization. To correctly handle data dependencies, if a reduction is the first operator in a fusion kernel, it is implemented with two loops: the first computes the reduction and the second uses it. Otherwise, each kernel is implemented as a single loop.

Reduction operations in statistical normalizations use a warp allreduce among all threads in a warp, implemented with CUDA intrinsics. If the number of elements to be reduced exceeds the warp size, we perform a sequential reduction over smaller chunks first. Layout-permuting, we use vectorized loads, stores, and arithmetic within a single thread, and fall back to word-wise implementations otherwise. For dropout operators, which must generate a random mask, we use cuRAND for random number generation.

After fusion, we have the following fused element-wise and normalization operations. Fig.~\ref{fig:fusion-examples} illustrates several cases.\vspace{-0.5em}
\begin{itemize}[itemsep=0.1mm,parsep=0pt]
\item \textsc{aib}: Attention input bias.
\item \textsc{baob}: Backward attention output bias.
\item \textsc{baib}: Backward attention input bias.
\item \textsc{sm}: Softmax with scaling and dropout.
\item \textsc{brd}: Bias, ReLU, and dropout.
\item \textsc{bdrln}: Bias, dropout, residual, and layernorm.
\item \textsc{bsb}: Backward layernorm scale and bias.
\item \textsc{blnrd}: Backward layernorm dX and dropout, saving the intermediate result for the residual connection.
\item \textsc{bdrb}: Backward dropout, ReLU, and bias.
\item \textsc{ebsb}: Backward residual and layernorm scale and bias.
\item \textsc{bs}: Backward dropout and softmax with scaling.
\item \textsc{bei}: Backward encoder input residual connection.
\end{itemize}\vspace{-0.5em}

\subsection{Evaluation Details}
\label{subsec:supp-eval-details}

All our results were produced on the Lassen supercomputer~\cite{lassen}, which consists of 795 nodes, each with two IBM Power9 CPUs and four Nvidia V100 GPUs with NVLINK2 and 16 GB of memory. We use CUDA 10.1.243 and GCC 7.3.1 to build our code. For comparison, we use cuDNN 7.6.5, PyTorch 1.6.0 (PT) (built-in transformer implementation), TensorFlow 2.1 from IBM PowerAI with XLA enabled (transformer adapted from~\citet{wolf2019huggingfacests}) (TF+XLA), and a recent development version of DeepSpeed (DS). Unless noted, our results use mixed-precision training~\cite{micikevicius2018mixed}, with FP16 data and accumulations performed at FP32. In PyTorch, we use Apex~\cite{apex} for mixed-precision; in TensorFlow and DeepSpeed, we use the built-in automatic mixed-precision. Results are the mean of 100 measurements. When we compute the percent of peak performance, we use the 125 Gflop/s Tensor Core peak on our V100s for tensor contraction operations, and the 31.4 Gflop/s half-precision peak for other operations.

\subsection{Flop Analysis}
\label{subsec:supp-flop-analysis}

Tab.~\ref{tab:flop-analysis} presents our complete flop analysis. See Sec.~\ref{subsec:fusion-results} for details.

\begin{table*}\footnotesize
  \vspace{-0.2cm}
  \caption{Flop analysis for BERT encoder layer. \tenscons{} -- tensor contraction, \statnorm{} -- statistical normalization, \elewise{} -- element-wise. MHA operators are filled black. We bold whichever is greater, \% peak (compute-bound) or MUE (memory-bound). The speedup is computed for kernels in isolation, overall speedup may be different due to measurement overheads.}
  \label{tab:flop-analysis}
  \centering
  \setlength{\tabcolsep}{3pt}
  \begin{tabular}{llrrrrrrrrrrl}
\toprule
&                                  &         & Input      & Output      & \multicolumn{3}{c}{PyTorch}           & \multicolumn{3}{c}{Ours}                                          &              &     \\
\cmidrule(lr){6-8} \cmidrule(lr){9-11}                                                                                                                           
& Operator                         & Gflop   & (1e6)      & (1e6)       & Gflop    & Time ($\mu$s)    & \% peak & Time ($\mu$s) & \% peak                    & MUE                  & Speedup      & Kernel  \\
\midrule                                                                                                                                                                        
\multirow{19}{*}{\rotatebox{90}{Forward}}                                                                                                                                       
& \atttenscons{} Q, K, V           & 25.770  & 7.34       & 12.58       & 25.782   &   333            & 56.2    &   306         &  \textbf{61.2}             & 12                   & 1.08         & --- \\
& \attelewise{} Input bias         & 0.013   & 12.59      & 12.58       & 0.025    &    90            & 0.4     &    66         &        0.5                 & \textbf{78}          & 1.35         & \rdelim\}{1}{*}[\textsc{aib}] \\
& \atttenscons{} $QK^T$            & 4.295   & 8.39       & 33.55       & 4.329    &   189            & 16.5    &   143         &        21.8                & \textbf{50}          & 1.32         & --- \\
& \attstatnorm{} Scaled softmax    & 0.201   & 33.55      & 100.66      & 0.956    &   453            & 1.3     &   433         &        1.3                 & \textbf{32}          & 1.04         & \rdelim\}{1}{*}[\textsc{sm}] \\
& \atttenscons{} Gamma             & 4.295   & 37.75      & 4.19        & 8.598    &   142            & 21.9    &   160         &        \textbf{19.4}       & 6                    & 0.88         & --- \\
& \atttenscons{} Out               & 8.590   & 5.24       & 4.19        & 8.686    &   136            & 45.9    &   120         &        \textbf{52}         & 10                   & 1.13         & --- \\
& \attelewise{} Output bias        & 0.004   & 4.20       & 4.19        & 0.008    &    34            & 0.4     & \mr{4}{  102} & \mr{4}{0.1}                & \mr{4}{\textbf{42}}  & \mr{4}{1.68} & \rdelim\}{4}{*}[\textsc{drln}] \\
& \elewise{} Dropout               & 0.004   & 4.19       & 8.39        & 0.014    &    37            & 0.3     &               &                            &                      &              & \\
& \elewise{} Residual              & 0.004   & 8.39       & 4.19        & 0.008    &    36            & 0.3     &               &                            &                      &              & \\
& \statnorm{} LayerNorm            & 0.029   & 4.20       & 4.19        & 0.052    &    63            & 1.3     &               &                            &                      &              & \\
& \tenscons{} Linear               & 34.360  & 8.39       & 16.78       & 34.377   &   451            & 55.4    &   402         &        \textbf{62.1}       & 12                   & 1.12         & ---\\
& \elewise{} Bias                  & 0.017   & 16.78      & 16.78       & 0.034    &   116            & 0.4     & \mr{3}{  183} & \mr{3}{0.3}                & \mr{3}{\textbf{76}}  & \mr{3}{1.90} & \rdelim\}{3}{*}[\textsc{brd}] \\
& \elewise{} ReLU                  & ---     & 16.78      & 16.78       & ---      &   112            & 0       &               &                            &                      &              & \\
& \elewise{} Dropout               & 0.017   & 16.78      & 33.55       & 0.052    &   120            & 0.4     &               &                            &                      &              & \\
& \tenscons{} Linear               & 34.360  & 20.97      & 4.19        & 34.456   &   449            & 55.6    &   369         &        \textbf{67.6}       & 6                    & 1.21         & --- \\
& \elewise{} Bias                  & 0.004   & 4.20       & 4.19        & 0.008    &    35            & 0.3     & \mr{4}{  101} & \mr{4}{0.1}                & \mr{4}{\textbf{43}}  & \mr{4}{1.70} & \rdelim\}{4}{*}[\textsc{bdrln}] \\
& \elewise{} Dropout               & 0.004   & 4.19       & 8.39        & 0.014    &    37            & 0.3     &               &                            &                      &              & \\
& \elewise{} Residual              & 0.004   & 8.39       & 4.19        & 0.008    &    37            & 0.3     &               &                            &                      &              & \\
& \statnorm{} LayerNorm            & 0.029   & 8.39       & 4.19        & 0.052    &    63            & 1.3     &               &                            &                      &              & \\
\midrule                                                                                                                                 
\multirow{27}{*}{\rotatebox{90}{Backward}}                                                                                                  
& \statnorm{} LayerNorm dW         & 0.017   & 8.39       & <0.01       & 0.021    &   184            & 0.3     &   150         &        0.3                 & \textbf{6}           & 1.22         & \rdelim\}{1}{*}[\textsc{bsb}] \\
& \statnorm{} LayerNorm dX         & 0.038   & 8.40       & 4.19        & 0.064    &    78            & 1.4     & \mr{2}{   71} & \mr{2}{1.5}                & \mr{2}{\textbf{37}}  & \mr{2}{1.58} & \rdelim\}{2}{*}[\textsc{blnrd}] \\
& \elewise{} Dropout dX            & 0.004   & 8.39       & 4.19        & 0.008    &    34            & 0.4     &               &                            &                      &              & \\
& \tenscons{} Linear+Bias dX       & 34.360  & 8.39       & 16.78       & 34.377   &   427            & 58.4    &   414         &        \textbf{60.3}       & 5                    & 1.03         & --- \\
& \tenscons{} Linear dW            & 34.360  & 20.97      & 4.19        & 34.389   &   424            & 58.9    &   378         &        \textbf{66}         & 13                   & 1.11         & --- \\
& \statnorm{} Bias dW              & 0.004   & 4.19       & <0.01       & 0.005    &    24            & 0.5     & \mr{4}{  362} & \mr{4}{<0.1}               & \mr{4}{\textbf{38}}  & \mr{4}{1.05} & \rdelim\}{4}{*}[\textsc{bdrb}] \\
& \elewise{} Dropout dX            & 0.017   & 33.55      & 16.78       & 0.034    &   129            & 0.4     &               &                            &                      &              & \\
& \elewise{} ReLU dX               & ---     & 33.55      & 16.78       & ---      &   166            & 0       &               &                            &                      &              & \\
& \statnorm{} Bias dW              & 0.017   & 16.78      & <0.01       & 0.021    &    61            & 0.8     &               &                            &                      &              & \\
& \tenscons{} Linear+Bias dX       & 34.360  & 20.97      & 4.19        & 34.389   &   417            & 59.9    &   398         &        \textbf{62.7}       & 6                    & 1.04         & --- \\
& \tenscons{} Linear dW            & 34.360  & 20.97      & 4.19        & 34.456   &   437            & 57.2    &   372         &        \textbf{67.2}       & 6                    & 1.17         & --- \\
& \elewise{} Residual              & 0.004   & 8.39       & 4.19        & 0.008    &    36            & 0.3     & \mr{2}{  250} & \mr{2}{<0.1}               & \mr{2}{\textbf{17}}  & \mr{2}{0.89} & \rdelim\}{2}{*}[\textsc{ebsb}] \\
& \statnorm{} LayerNorm dW         & 0.017   & 8.39       & <0.01       & 0.021    &   186            & 0.3     &               &                            &                      &              & \\
& \statnorm{} LayerNorm dX         & 0.038   & 8.40       & 4.19        & 0.064    &    80            & 1.4     & \mr{2}{   69} & \mr{2}{1.6}                & \mr{2}{\textbf{37}}  & \mr{2}{1.64} & \rdelim\}{2}{*}[\textsc{blnrd}] \\
& \elewise{} Dropout dX            & 0.004   & 8.39       & 4.19        & 0.008    &    34            & 0.4     &               &                            &                      &              & \\
& \attstatnorm{} Output bias dW    & 0.004   & 4.19       & <0.01       & 0.005    &    23            & 0.5     &    38         &        0.3                 & \textbf{22}          & 0.60         & \rdelim\}{1}{*}[\textsc{baob}] \\
& \atttenscons{} Out dX            & 8.590   & 4.33       & 4.19        & 8.637    &   131            & 47.6    &   119         &        \textbf{52.2}       & 10                   & 1.09         & --- \\
& \atttenscons{} Out dW            & 8.590   & 8.39       & 1.05        & 8.686    &   136            & 45.9    &   113         &        \textbf{54.8}       & 5                    & 1.19         & --- \\
& \atttenscons{} Gamma dX1         & 4.295   & 8.39       & 33.55       & 8.598    &   136            & 22.8    &   147         &        \textbf{21.2}       & 7                    & 0.93         & --- \\
& \atttenscons{} Gamma dX2         & 4.295   & 67.11      & 33.55       & 4.329    &   188            & 16.6    &   123         &        \textbf{25.2}       & 8                    & 1.52         & --- \\
& \attstatnorm{} Scaled softmax dX & 0.168   & 12.58      & 4.19        & 0.214    &   790            & 0.6     &   426         &        1.1                 & \textbf{49}          & 1.85         & \rdelim\}{1}{*}[\textsc{bs}] \\
& \atttenscons{} $QK^T$ dX1        & 4.295   & 37.75      & 4.19        & 4.299    &   135            & 23.1    &   155         &        \textbf{20}         & 7                    & 0.86         & --- \\
& \atttenscons{} $QK^T$ dX2        & 4.295   & 37.75      & 4.19        & 8.598    &   139            & 22.3    &   115         &        \textbf{26.9}       & 9                    & 1.20         & --- \\
& \atttenscons{} Q, K, V dX        & 25.770  & 15.73      & 4.19        & 25.799   &   344            & 54.4    &   274         &        \textbf{68.2}       & 6                    & 1.25         & --- \\
& \atttenscons{} Q, K, V dW        & 25.770  & 20.46      & 1.05        & 25.911   &   329            & 57      &   293         &        \textbf{64}         & 6                    & 1.12         & --- \\
& \attstatnorm{} Input bias dW     & 0.013   & 12.58      & <0.01       & 0.016    &    52            & 0.7     &    39         &        0.9                 &         \textbf{66}  & 1.32         & \rdelim\}{1}{*}[\textsc{baib}] \\
& \elewise{} Residual              & 0.004   & 8.39       & 4.19        & 0.008    &    35            & 0.3     &    31         &        0.4                 &         \textbf{83}  & 1.14         & \rdelim\}{1}{*}[\textsc{bei}] \\
\midrule                                                                                                                                                        
& \tenscons{} Tensor contractions  & 335.007 & ---        & ---         & 348.698  &  4951            & 43.1    &  4411         &        48.5                & ---                  & 1.12         & \\
& \statnorm{} Stat. normalizations & 0.575   & ---        & ---         & 1.492    &  2063            & 0.9     &  1591         &        0.6                 & ---                  & 1.29         & \\
& \elewise{} Element-wise          & 0.105   & ---        & ---         & 0.239    &  1096            & 0.3     &   735         &        0.1                 & ---                  & 1.49         & \\
& \opsymbspace{} Total             & 335.687 & ---        & ---         & 350.429  &  8110            & 31.1    &  6739         &        35                  & ---                  & 1.20         & \\
\bottomrule
  \end{tabular}
  \vspace{-1em}
\end{table*}

\subsection{Additional Fusion Opportunities}
\label{subsec:supp-fusion}

\begin{table}
  \caption{Algebraic fusion for MHA Q/K/V ($\mu$s).}
  \label{tab:fusion-algebraic}
  \centering
  \small
  \begin{tabular}{llll}
    \toprule
    & Unfused & $QK$ fused & $QKV$ fused \\
    \midrule
    Forward & 345 &  294 & \textbf{275} \\
    Backward & 342 &  312 & \textbf{291} \\
    \bottomrule
  \end{tabular}
  \vspace{-1em}
\end{table}

\subsubsection{Fusing into Tensor Contractions}
\label{subsec:fusion-fuse-tc}

Because cuBLAS does not support fusing arbitrary operators into (batched) MMMs, we evaluated CUTLASS~\cite{cutlass} version 2.1 as an alternative, which does support fusing element-wise operators. We conducted a simple benchmark comparing cuBLAS with a separate bias kernel to CUTLASS for the first linear layer of BERT. We found that the batched MMM in CUTLASS is approximately 40 $\mu$s slower than cuBLAS. The reduction in runtime by fusing the bias is less than this. Hence, we only consider cuBLAS for tensor contractions. cuBLAS does support simple scaling operations, which we use to implement the scaling for the scaled softmax operator.

\subsubsection{Algebraic Fusion}
\label{subsec:fusion-algebraic}

There is an additional opportunity for fusion among certain tensor contraction operators. Using domain knowledge and the dataflow graph, we can identify some operations that can be combined into a single algebraically equivalent operation. For example, there are several different ways to perform the input projections (batched MMMs) in self-attention, since the input queries, keys, and values are the same tensor, $X$:
\begin{enumerate}
    \item $X$ can be multiplied by each of the projection matrices: $W^Q X$, $W^K X$, and $W^V X$.
    \item The $W^Q$ and $W^K$ matrices can be stacked and two multiplications performed: $[W^Q \: W^K] X$ and $W^V X$. Similarly, the $W^K$ and $W^V$ matrices can be stacked.
    \item All three can be stacked: $[\widetilde Q \: \widetilde K \: \widetilde V] = [W^Q \: W^K \: W^V] X$.
\end{enumerate}
In backpropagation, the $dW$ and $dX$ computations for each projection matrix can be similarly fused: $X [d \widetilde Q \: d \widetilde K \: d \widetilde V]$ and $[W^Q \: W^K \: W^V] [d \widetilde Q \: d \widetilde K \: d \widetilde V]$, respectively.

There are different tradeoffs to these approaches, which must be determined empirically. Performing separate MMMs may enable task parallelism. On the other hand, this stacking enables data reuse, since $X$ is used only once.

Tab.~\ref{tab:fusion-algebraic} presents results for each of these cases. Fully fusing this batched MMM performs best. Unfortunately, cuBLAS launches kernels that utilize the entire GPU, so task parallelism is not profitable. This example can also be adapted to fuse keys and values in encoder/decoder attention.

\subsection{End-to-end training}
\label{subsec:supp-e2e}

We conducted end-to-end training of BERT-large on the Wikipedia data corpus. We adapted the BERT implementation from the Nvidia deep learning examples~\cite{nvdlex} and performed training on Lassen using 32 nodes with 128 GPUs. The training consists of two phases, one using a sequence length of 128 and the other with a sequence length of 512.

\begin{figure}
  \centering
  \includegraphics[width=\linewidth]{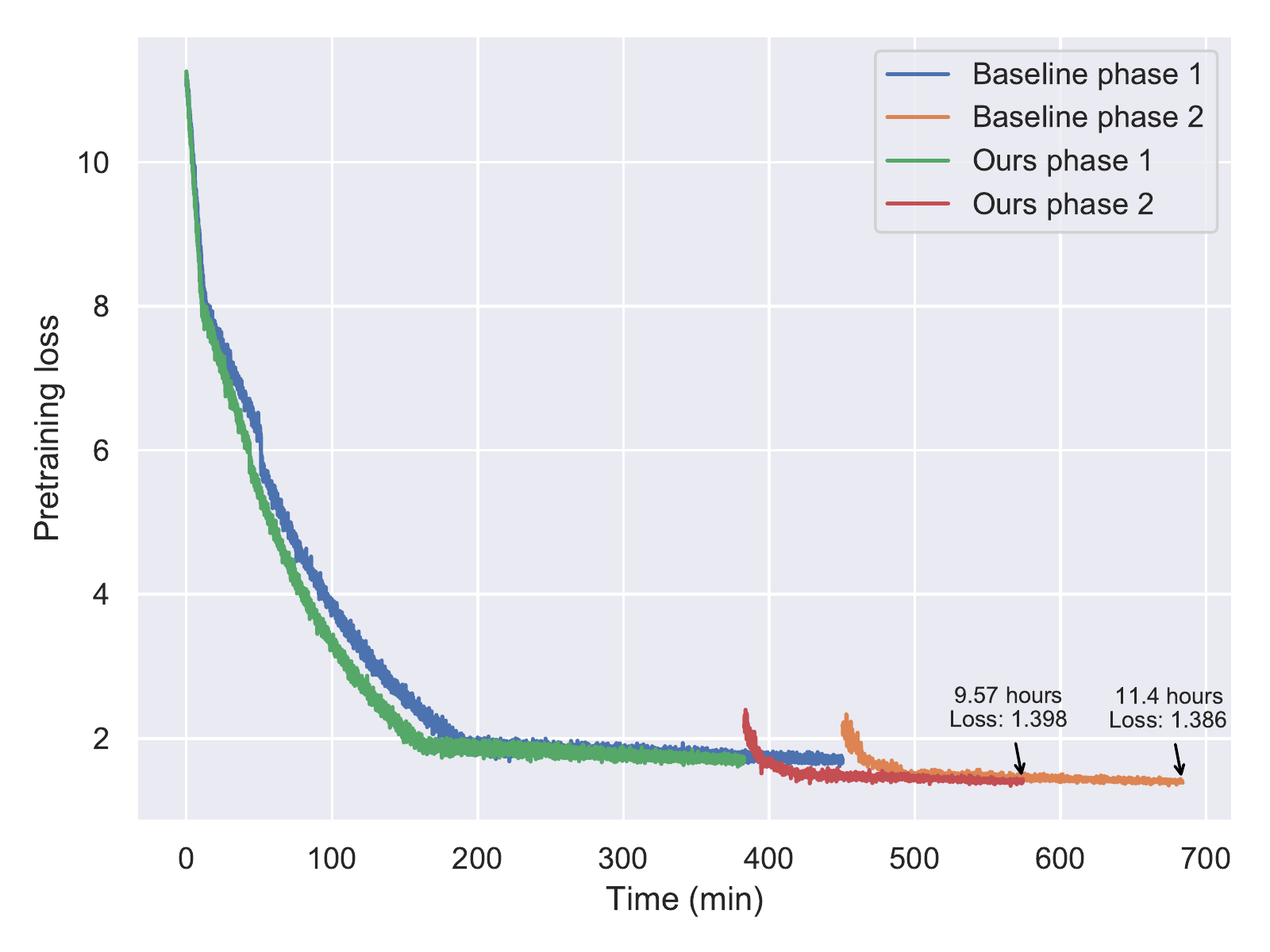}
  \caption{Pretraining loss and time curves for BERT.}
  \label{fig:supp-pretraining}
\end{figure}

Pretraining loss versus time is presented in Fig.~\ref{fig:supp-pretraining}. We achieve a $1.18\times$ speedup in phase 1 and a $1.22\times$ speedup in phase 2, for an overall speedup of $1.19\times$.
Note this is less than our $1.30\times$ speedup over PyTorch for an encoder layer for several reasons: The Nvidia implementation includes some further optimizations, such as fused layer normalization kernels; there are additional operations (e.g., embedding layers, loss functions) that we do not currently optimize; and overheads from distributed training and data loading.

\subsection{Additional Related Work}
\label{subsec:supp-related}

There has been significant work optimizing both transformers in particular and deep learning in general. For a comprehensive overview, see~\citet{ddlsurvey}. To help guide training regimes for transformers, recent work has provided empirical recommendations on model size, batch size, and so on~\cite{li2020train,kaplan2020scaling,rosenfeld2019constructive}. Many of the subsequent techniques we review are complementary to our work.

Most directly relevant are other approaches specifically to accelerate transformer training. Distributed-memory techniques, such as ZeRO~\cite{rajbhandari2019zero}, Megatron~\cite{shoeybi2019megatron}, and Mesh-TensorFlow~\cite{shazeer2018mesh} scale training to many GPUs to accelerate it. Mesh-TensorFlow also presents a classification of operators similar to ours. Large batches have also been used to accelerate training via LAMB~\cite{you2019large} or NVLAMB~\cite{nvlamb}. None of these directly address the performance of a single GPU as done in this paper. DeepSpeed~\cite{deepspeed}, which we compare with in Section~\ref{subsec:e2e-e2e}, is closest to our work, but performs all optimizations and layout selections manually.


Transformer architectures to enable improved training have also been the subject of significant recent work. ALBERT~\cite{lan2019albert} used a combination of weight sharing and factorized embedding layers to reduce memory requirements; however compute times are unaffected. Transformer-XL~\cite{dai2019transformer} caches prior hidden states to learn long sequences. RevNets~\cite{gomez2017reversible}, a variant of ResNets which allow activations to be reconstructed during backpropagation, have been applied to transformers. Notably, Reformer~\cite{kitaev2020reformer} combines reversible residual layers with locality-sensitive hashing to improve the efficiency of multi-head attention. Sparsity optimizations for attention~\cite{bahdanau2014neural,luong2015effective,shen2018bi,parmar2018image,tay2019simple,child2019generating,correia2019adaptively,sukhbaatar2019adaptive,tay2020sparse} reduce memory and compute requirements. We view these as orthongal to our work: the same principles of data-flow analysis can be applied to optimize for sparsity and reuse.

There has also been much work on optimizing deep learning in general. Many frameworks provide implementations of transformers or their components, such as PyTorch~\cite{paszke2019pytorch}, TensorFlow~\cite{tensorflow2015-whitepaper}, cuDNN~\cite{chetlur2014cudnn}, and others built atop these~\cite{ott2019fairseq,wolf2019huggingfacests}. Optimizing frameworks can also be applied to transformers~\cite{xla,frostig2018compiling,jax2018github,torchscript,rotem2018glow,jia2019beyond,jia2019optimizing,jia2019taso,chen2018tvm,nnvm,sivathanu2019astra,mirhoseini2017device,cyphers2018intel,baghdadi2019tiramisu,vasilache2018tensor,lethin2019polyhedral,wei2017dlvm,truong2016latte,venkat2019swirl,dong2019acorns,elango2018diesel}. None of these frameworks provide all the optimizations or the systematic study of data movement and its impact on performance. We have specifically compared against some of the most popular production frameworks: PyTorch, TensorFlow with XLA, and cuDNN. Beyond these, TASO~\cite{jia2019taso} targets similar optimizations to ours by using graph substitutions, but considers only inference and does not exhaustively explore the search space.


Other optimizations, including model parallelism~\cite{van2015lbann,jia2019beyond,gholami2018integrated,dean2012large,chilimbi2014project,shazeer2018mesh,buchlovsky2019tf}, pipeline parallelism~\cite{chen2012pipelined,li2018pipe,narayanan2019pipedream,huang2019gpipe}, microbatching~\cite{oyama2018accelerating}, and recomputation for memory reduction~\cite{chen2016training,jain2019checkmate} are all also applicable. Communication can also be a major bottleneck for training transformers, due to the large model size~\cite{shoeybi2019megatron,shazeer2018mesh}. Frameworks for inference, including TensorRT~\cite{tensorrt}, Caffe2~\cite{caffe2}, and the ONNX Runtime~\cite{onnxruntime}, all help to enable a suite of optimizations primarily applicable during inference. Pruning~\cite{shazeer2019fast,voita2019analyzing} and distillation~\cite{sanh2019distilbert} has also been used to accelerate inference.

Neural architecture-specific optimizations have a long history outside of transformers, and have primarily targeted CNNs~\cite{krizhevsky2012imagenet,coates2013deep,goyal2017accurate,akiba2017extremely,you2018imagenet,mikami2018imagenet,ying2018image,dryden2019improving,dryden2019channel}. Notably, \citet{li2016optimizing} optimized data layouts for convolution.

In general, data movement reduction is a core component of high-level optimization~\cite{tpds-locality}. Optimizing compilers, most notably components that specialize in polyhedral programs~\cite{polly,pluto}, apply loop transformations (e.g., tiling, skewing) that belong to the class of data movement optimization. Other white-box approaches for separation of program definition from data optimization passes include Halide~\cite{ragan2013halide}, JAX~\cite{frostig2018compiling,jax2018github}, Legion~\cite{bauer2012legion}, Lift~\cite{lift}, and MLIR~\cite{lattner2020mlir}. The data-centric approach proposed here enables user-extensible coarse- and fine-grained data movement optimization via the flat (yet parametric) dataflow graph representation. This allows us to perform and tune complex data layout and fusion transformations that span multiple granularity levels, surpassing the optimization capabilities of the aforementioned tools and achieving state-of-the-art performance.


\end{document}